\documentclass{article}

\usepackage[preprint]{template_style}

\usepackage[utf8]{inputenc} 
\usepackage[T1]{fontenc}    
\usepackage{url}            
\usepackage{booktabs}       
\usepackage{amsfonts}       
\usepackage{nicefrac}       
\usepackage{microtype}      
\usepackage{xcolor}         
\usepackage{graphicx}
\usepackage{xspace}
\usepackage{amsmath}
\usepackage{float}
\usepackage[pagebackref=true,breaklinks,colorlinks,citecolor=blue,linkcolor=blue,urlcolor=blue,hypertexnames=false]{hyperref}

\title{Low-Rank Head Avatar Personalization \\ with Registers}

\author{%
  Sai Tanmay Reddy Chakkera \\
  Department of Computer Science\\
  Stony Brook University\\
  \texttt{schakkera@cs.stonybrook.edu} \\
  \And
  Aggelina Chatziagapi \\
  Department of Computer Science \\
  Stony Brook University\\
  \texttt{echatziagapi@cs.stonybrook.edu} \\
  \And
  Md Moniruzzaman \\
  Atmanity Inc. \\
  \texttt{mman@atmanity.io} \\
  \And
  Chen-Ping Yu \\
  Atmanity Inc. \\
  \texttt{cpyu@atmanity.io} \\
  \And
  Yi-Hsuan Tsai \\
  Atmanity Inc. \\
  \texttt{yhtsai@atmanity.io} \\
  \And
  Dimitris Samaras \\
  Department of Computer Science \\ 
  Stony Brook University \\ 
  \texttt{samaras@cs.stonybrook.edu}\\
}
\newcommand{\ModuleName}{Register Module\xspace}

\begin{document}

\maketitle
\begin{abstract}
  We introduce a novel method for low-rank personalization of a generic model for head avatar generation. Prior work proposes generic models that achieve high-quality face animation by leveraging large-scale datasets of multiple identities. However, such generic models usually fail to synthesize unique identity-specific details, since they learn a general domain prior.
  To adapt to specific subjects, we find that it is still challenging to capture high-frequency facial details via popular solutions like low-rank adaptation (LoRA).
  This motivates us to propose a specific architecture, a Register Module, that enhances the performance of LoRA, while requiring only a small number of parameters to adapt to an unseen identity. Our module is applied to intermediate features of a pre-trained model, storing and re-purposing information in a learnable 3D feature space. To demonstrate the efficacy of our personalization method, we collect a dataset of talking videos of individuals with distinctive facial details, such as wrinkles and tattoos. Our approach faithfully captures unseen faces, outperforming existing methods quantitatively and qualitatively. We will release the code, models, and dataset to the public. Project page: \href{https://starc52.github.io/publications/2025-05-28-LoRAvatar/}{https://starc52.github.io/publications/LoRAvatar/}.
\end{abstract}
\section{Introduction}
\label{sec:intro}

Synthesizing photo-realistic human faces has long been a challenge for both computer vision and graphics. It has broad applications from AR/VR, virtual communication, and video games, to the movie industry and healthcare. Earlier approaches rely on 3D morphable models (3DMMs)~\citep{garrido2015vdub,garrido2014automatic,thies2016face2face}, while subsequent methods turn to generative adversarial networks (GANs)~\citep{kim2018deepvideo,pumarola2020ganimation,wav2lip,vougioukas2020realistic}.
More recent works learn 3D neural representations of the human face, which rely on neural radiance fields (NeRFs)~\citep{pumarola2021d,nerfies,nerface,park2021hypernerf} or 3D Gaussian Splatting (3DGS)~\citep{kerbl20233d,cho2024gaussiantalker,qian2024gaussianavatars,xu2023gaussianheadavatar}. While these approaches lead to high-quality results, they usually require identity-specific training and are not able to generalize. Only a few recent methods propose \textit{generic} models, e.g., GAGAvatar~\citep{chu2024gagavatar}, which preserve the high-quality rendering of 3DGS, while trained on a large-scale dataset of multiple identities, enabling generalization to unseen human faces.

However, such generic models usually fail to produce key identity-specific facial details, since they learn a general domain prior. To produce distinctive details, prior work proposes adapting a pre-trained model to a specific identity, e.g.~through fine-tuning or meta-learning~\citep{nitzan2022mystyle,zhang2023metaportrait,saunders2024talklora}.
Low-rank adaptation (LoRA)~\citep{hu2022lora} has been first proposed for large language models (LLMs). It injects trainable rank decomposition matrices into each layer of a pre-trained model, leading to a significant decrease of the learnable parameters and on-par performance compared to fine-tuning the entire model.

\begin{figure}[!t]
    \centering
    \includegraphics[width=0.85\linewidth]{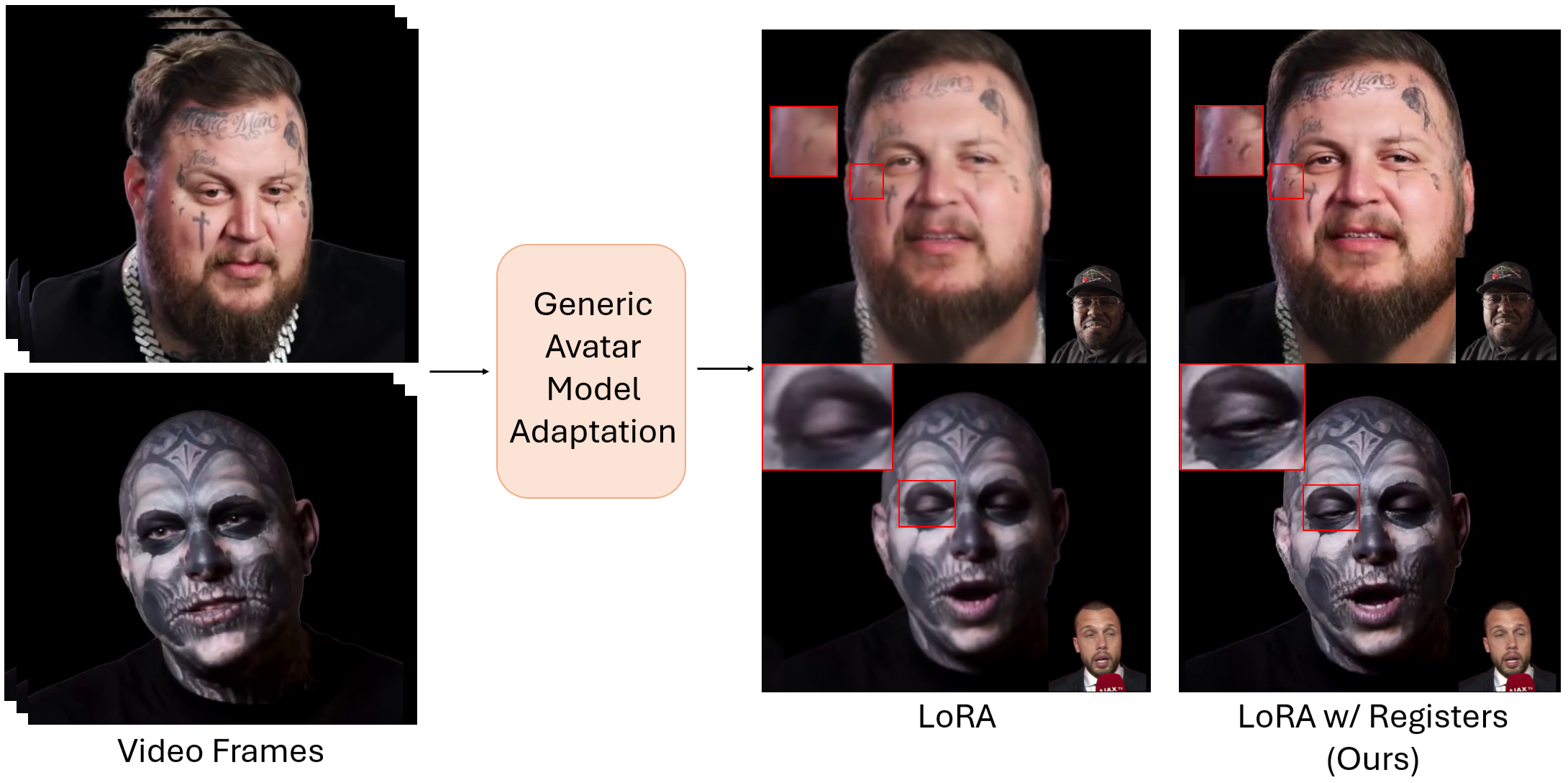}
    \caption{
    Our method personalizes and adapts a generic head avatar model using LoRA, while preserving high-frequency identity-specific facial details using our \ModuleName and retaining the original inference speed. Note that the small image in the bottom-right corner is the driving image.
    }
    \label{fig:teaser}
\end{figure}

In this work, we address the problem of adaptation, also called \textit{personalization}, to a specific identity, which is not seen in the initial training of a generic model for head avatar generation. Due to its efficiency and popularity in other fields, we start with LoRA, by learning low-rank decomposition matrices for specific layers. We notice that LoRA is not sufficient to synthesize high-frequency facial characteristics (see Figure~\ref{fig:teaser}). Inspired by~\cite{darcet2023vision} that learn additional tokens (registers) in order to store global information for a transformer network, we propose a specific module that extends the idea of registers to 3D registers for human faces. To the best of our knowledge, we believe that this is the first method to extend registers to 3D representations.

More specifically, we design a \ModuleName that learns a 3D feature space that stores and repurposes information for a human face during training. Similar to registers in ViT~\citep{darcet2023vision} that store global information of an image, our \ModuleName stores the distinctive details of an identity, given different views. We apply our \ModuleName to intermediate features that are extracted from a pre-trained DINOv2 model~\citep{oquab2023dinov2}. While our proposed module can be applied to any network that uses DINOv2 features, we focus our study on GAGAvatar~\citep{chu2024gagavatar} as our generic pre-trained model. To evaluate the efficacy of our low-rank personalization, we collect a dataset of talking videos of individuals with rare high-frequency facial details, such as wrinkles and tattoos, that are not included in existing datasets.
Our method outperforms state-of-the-art approaches, like meta-learning and vanilla LoRA, both quantitatively and qualitatively, while it only requires a small number of parameters to adapt.

In brief, our main contributions are as follows:
\begin{itemize}
    \item We propose a novel method for low-rank personalization of a generic model for head avatar generation, that captures identity-specific facial details.
    \item We design a \ModuleName that stores and repurposes information for an identity in a learnable 3D feature space, extending the idea of registers for ViTs to 3D human faces.
    \item We collect a dataset, namely RareFace-50, of talking videos of individuals with distinctive facial characteristics, e.g.~wrinkles and tattoos, that are challenging to synthesize with generic models, and thus demonstrating the need for our method.
\end{itemize}

\section{Related Work}
\label{sec:related_work}

\noindent
\textbf{Human Portrait Synthesis.}
Earlier approaches for video synthesis of human faces are based on 3DMMs~\citep{garrido2015vdub,garrido2014automatic,thies2016face2face}. A 3DMM~\citep{blanz1999morphable} is a parametric model that represents a face as a linear combination of the principal axes of shape, texture, and expression, learned by principal component analysis (PCA). Subsequent works propose GAN-based networks for video synthesis~\citep{kim2018deepvideo,fomm,pumarola2020ganimation} and audio-driven talking faces~\citep{wav2lip,pcavs,vougioukas2020realistic,xu2024emotion}. GANs are usually trained on large datasets of 2D videos of multiple identities, but they cannot model the 3D face geometry.
More recent works learn 3D neural representations of the human face, which rely on neural radiance fields (NeRFs)~\citep{mildenhall2020nerf} or 3D Gaussian Splatting (3DGS)~\citep{kerbl20233d}. Diffusion models have also become popular but they only produce 2D videos~\citep{xu2024vasa} or are identity-specific~\citep{kirschstein2024diffusionavatars}. 
In this paper, we explore personalization to capture identity-specific facial details by adapting a generic avatar model.

\noindent
\textbf{Animatable 3D Head Avatars.}
NeRFs have been first proposed for novel-view synthesis of static scenes~\citep{mildenhall2020nerf}. They have been extended to dynamic scenes and human faces~\citep{pumarola2021d,nerfies,nerface,park2021hypernerf,chakkera2024jean}. They usually represent a human face by sampling 3D points in a canonical space, which can be conditioned on 3DMM expression parameters to enable animation. Although they produce high-quality reconstructions, they require expensive identity-specific training.
Subsequent works~\citep{zielonka2023insta,bakedavatar} propose techniques to reduce the training and inference time. 3DGS~\citep{kerbl20233d} became very popular as it achieves real-time rendering with high visual quality, by representing complex scenes with 3D Gaussians. It has recently been applied for dynamic human avatars~\citep{cho2024gaussiantalker,qian2024gaussianavatars,xu2023gaussianheadavatar,dhamo2024headgas,wang2025gaussianhead}. However, most approaches learn identity-specific models. Very few recent works propose generic models~\citep{chu2024gagavatar,chu2024gpavatar,kirschstein2024gghead}, which preserve the high-quality rendering of 3DGS, while trained on a large-scale dataset of multiple identities, enabling generalization to unseen human faces.
However, generic models learn a general domain prior and usually fail to produce unique identity-specific facial details, such as wrinkles or tattoos, as studied in this paper.

\noindent
\textbf{Personalization.}
Numerous works have proposed ways to adapt pre-trained models to various downstream tasks~\citep{houlsby2019parameter,zhang2023adding}. Parameter-efficient fine-tuning (PEFT) techniques
are proposed to fine-tune large models efficiently. LoRA~\citep{hu2022lora} adds low-rank matrices into each layer of a pre-trained model, leading to a significant decrease of the learnable parameters and on-par performance compared to fine-tuning the entire model. In the context of face animation, fine-tuning part of the model has been utilized~\citep{chatziagapi2024mi,li2025instag}, as well as  meta-learning. For instance, MetaPortrait~\citep{zhang2023metaportrait} adopts a meta-learning approach to allow adaptation during inference, while \cite{gao2020portrait} uses meta-learning to adapt a NeRF to a single image of an unseen subject. Moreover, MyStyle~\citep{nitzan2022mystyle} personalizes a pre-trained StyleGAN by fine-tuning regions of its latent space, using a set of images from an individual.
Similarly, One2Avatar~\citep{yu2024one2avatar} adapts a generic NeRF to one or a few images of a person.
TalkLoRA~\citep{saunders2024talklora} applies LoRA for the task of 3D mesh animation, while My3DGen~\citep{qi2025my3dgen} adapts LoRA to the convolutional layers of StyleGAN2 in an EG3D-based network~\citep{chan2022efficient}.
Due to its popularity and efficiency, we study LoRA for our generic model for head avatar animation. However, we find that LoRA is not sufficient to capture high-frequency facial details of a new identity. Thus, we propose to improve personalization by learning an additional register module, inspired by register tokens in ViTs.

\noindent
\textbf{Additional Tokens in Neural Networks.}
Memory augmentation in neural networks goes back to long short-term memory (LSTM) units~\citep{hochreiter1997long} that store information through gates. Memory networks~\citep{weston2014memory,sukhbaatar2015end} have access to external long-term memory. More recently, transformers have emerged as a powerful representation for various deep learning tasks, where the core element is self-attention~\citep{vaswani2017attention}. For language modeling, many works extend the input sequence of transformers with special tokens. Such additional tokens provide the network with new information, e.g.~[SEP] in BERT~\citep{devlin-etal-2019-bert}, or gather information for later downstream tasks, e.g.~[CLS] tokens~\citep{dosovitskiy2021an}, or [MASK] for generative modeling~\citep{bao2021beit}. Unlike these works, \cite{darcet2023vision} present additional tokens as registers for storing and repurposing global information. Inspired by this, we extend registers to a 3D feature space for human faces. We learn a \ModuleName that stores information about distinctive high-frequency details of a human face.
\section{Proposed Method}
\label{sec:method}
\begin{figure}
    \centering
    \includegraphics[width=\linewidth]{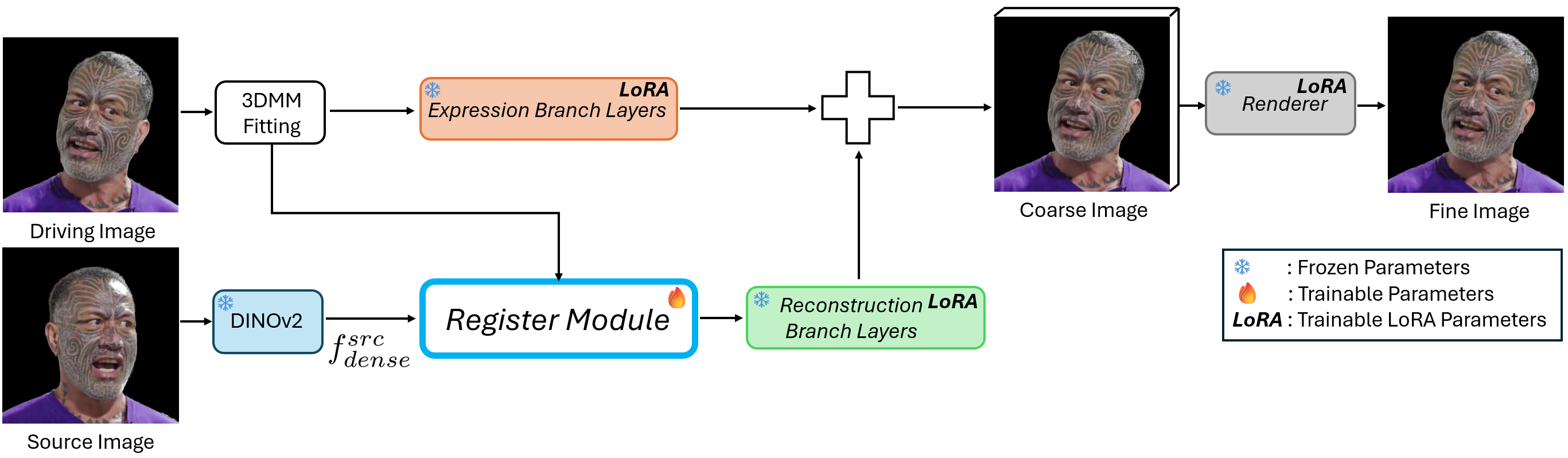}
    \caption{Illustration of our \ModuleName in a generic avatar animation model. During adaptation, we pass the source image's DINOv2 features $f^{src}_{dense}$ and driving image's 3DMM parameters to our module. Our module teaches the model to attend to specific regions in the dense DINOv2 features, thus providing better learning signals for LoRA to capture identity-specific details. Note that the \ModuleName is not needed during inference but serves as the register during LoRA training.}
    \label{fig:full_figure}
        \vspace{-15pt}
\end{figure}
Figure~\ref{fig:full_figure} illustrates an overview of our proposed framework that adapts a general avatar model to a particular identity. Inspired by Parameter-Efficient Fine-Tuning (PEFT), specifically LoRA \citep{hu2022lora}, we utilize LoRA to adapt the weights of a generalized avatar model to a particular identity. Through experiments, initially we find that adapting with LoRA does not sufficiently improve personalization (see Figure~\ref{fig:teaser}). Motivated by \cite{darcet2023vision} that introduce registers in ViTs to store and repurpose global information, we propose a Register Module to store information about identity-specific details. Our \ModuleName essentially teaches the model to attend to specific regions in the dense DINOv2 features during adaptation. Importantly, it is only used during adaptation and is deactivated at inference time. 
With its guidance, the model learns to leverage DINOv2 features more effectively, enabling high-quality personalized head avatar generation with real-time speed from a single source image at inference.

Our proposed pipeline for personalizing a generic avatar animation model consists of two main components:

(1) We add LoRA weights to specific pre-trained layers of a generic avatar animation model (see Sec.~\ref{sec:method_general_model}), to keep the adaptation parameters efficient and to avoid catastrophic forgetting (see Sec.~\ref{sec:method_lora}).

(2) We design a Register Module that learns a 3D feature space, facilitating the attention to specific regions of DINOv2 features, while adapting to a face from multiple views (see Sec.~\ref{sec:method_register_module}).

We first describe a generic avatar animation model in Sec.~\ref{sec:method_general_model}. Next, we describe the process of adding LoRA weights to pre-trained layers in Sec.~\ref{sec:method_lora}. Finally, we describe the architecture of our \ModuleName in Sec.~\ref{sec:method_register_module}.

\subsection{Preliminaries: Generic Avatar Generation}
\label{sec:method_general_model}
An avatar generation model consists of two branches: (a) reconstruction branch, and (b) expression branch. The reconstruction branch generates an animatable head avatar from the source image. The expression branch extracts the expressions and pose from the driving image which is used to animate the generated head avatar. These branches are merged and the output is rendered using a neural renderer. This process learns a model for generalized head avatar reconstruction (see Figure \ref{fig:full_figure}).

In particular, we use GAGAvatar~\citep{chu2024gagavatar} as our generic model trained on a large-scale dataset using the general DINOv2 features in its reconstruction branch, which is suitable to serve as our foundation model for fast adaptation.
While DINOv2 features are robust for generic tasks, they may contain irrelevant information for avatar generation. Thus, adapting the layers of the generic avatar model to focus on relevant information within the DINOv2 feature space is necessary. We propose our Register Module for this purpose in the following sections.

\subsection{LoRA for Fast Adaptation}
\label{sec:method_lora}
To adapt to a particular identity, inspired by the literature in NLP, we use LoRA~\citep{hu2022lora} for adaptation in a parameter-efficient manner. For a pre-trained weight matrix $W \in \mathbb{R}^{m \times n}$, LoRA models the adapted weights $W_{adapt}$ by representing it as an addition of the pretrained weights $W$ and an offset matrix $\Delta W$, the latter of which is low-rank decomposable.
\begin{equation}
    W_{adapt} = W + \Delta W = W + BA,
\end{equation}
where $B \in \mathbb{R}^{m \times r}$ and $A \in \mathbb{R}^{r \times n}$ and $r \ll min(m, n)$. During adaptation, only parameters in $A$ and $B$ receive gradients. 
For our purpose, we add LoRA weights $A$ and $B$ to each parameter matrix in a pre-trained avatar model, except the DINOv2 model. 
In our implementation, for all experiments except ablations described in the supplementary material, we use the same rank $r=32$ for all comparisons.

\subsection{Register Module}
\label{sec:method_register_module}
\begin{figure}
    \centering
    \includegraphics[width=\linewidth]{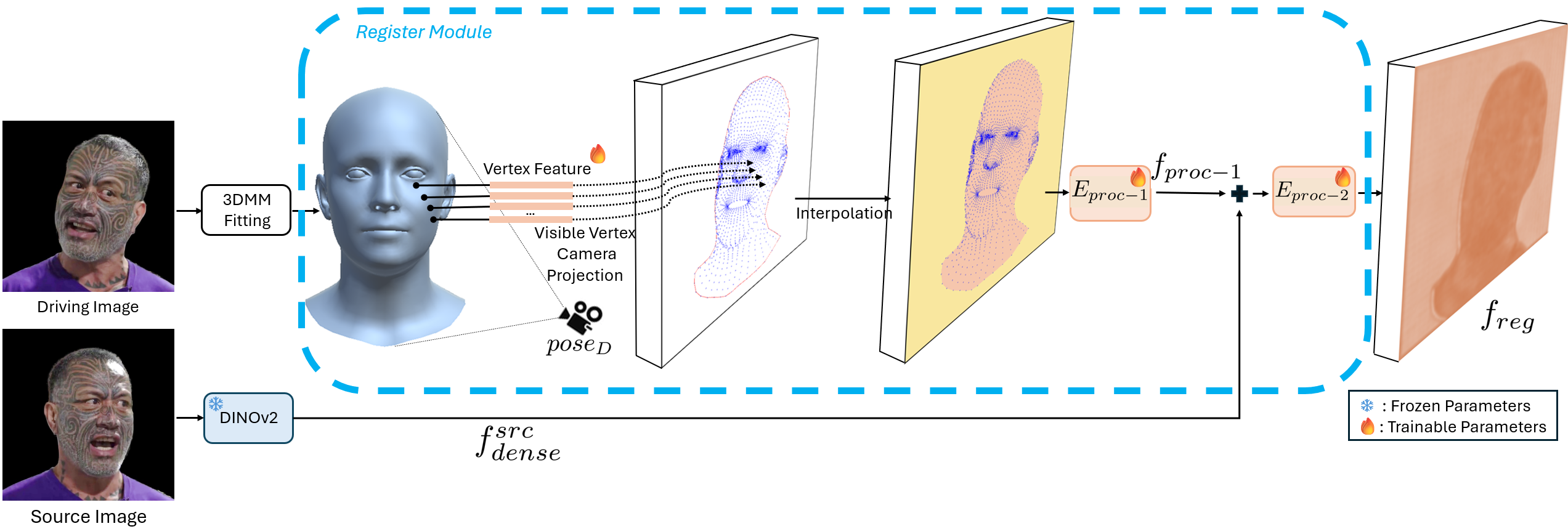}
    \caption{Illustration of our Register Module. We propose a Register Module that learns features in the 3D space. We rig embeddings to vertices on a 3DMM mesh and use camera pose $pose_D$ to project visible vertices and their embeddings onto a camera plane. Next, we interpolate features in the face mask region, fill in the background feature. Finally, we add these features to source image's DINOv2 features $f^{src}_{dense}$ to improve learning signals for LoRA.}
    \label{fig:main_diagram}
        \vspace{-15pt}
\end{figure}
Figure \ref{fig:main_diagram} illustrates the design of our \ModuleName. We hypothesize that, in addition to adding LoRA weights, we need a mechanism for better extraction of fine details of an identity, such as tattoos, wrinkles, muscular idiosyncrasies, and other personal features of an identity. To this end, we introduce a \ModuleName to improve the focus on identity-specific details.

\textbf{Feature Learning Procedure.}
Specifically, we propose to highlight detailed information in DINOv2 local features of the source image with the output of the \ModuleName. Let $M = (V, E, F)$ be a 3DMM mesh \citep{FLAME:SiggraphAsia2017}, where $V$ is the set of vertices, $n(V)$ is the number of vertices, $E$ is the set of edges and $F$ is the set of facets. In our \ModuleName, we rig embeddings $e \in \mathbb{R}^{n(V) \times D}$, where D is the dimension of the embeddings, to vertices $v \in V$ of mesh $M$. Given a driving image camera pose and position $pose_D$, we compute the set of visible vertices $U \subset V$ of the mesh $M$ from 
\begin{equation}
    U=\texttt{visible}(M, pose_D) \;.
\end{equation}
Next, we project these points $U$ to a feature space in the camera plane $S = \{(i,j) \in \mathbb{Z}^2 | 1 \leq i \leq H,1 \leq j \leq W\}$ and their corresponding embeddings to a dense feature $f_S \in \mathbb{R}^{H \times W \times D}$, where $H, W$ are the parameters of the image size, using a Perspective Projection: 
\begin{equation}
    U_{S}=\texttt{perspective\_project}(U, pose_D, K(S)) \;,
\end{equation}

where $K(S)$ is the intrinsic camera matrix for a camera with $S$ as the camera plane. The projected points are rounded off to the nearest integer. At points in the feature plane where a visible vertex $v \in U$ is projected to, we assign the corresponding point's embedding from $e$. Hence the operation becomes: 
\begin{equation}
    f_S[u^{i}_S] := e[u^{i}] \text{ for } u^i\in U \text{ and } u^i_{S} \text{ is the projection of } u^i \;.
\end{equation}

Given the set of points $U_s$ on the camera plane, we compute an alpha shape \citep{alpha_shape} to find the simple contour polygon $P_{U_S}$ of the vertex projections. Let $\texttt{interior}(P)$ represent all the points inside a simple polygon $P$. For each point $p \in \texttt{interior}(P_{U_S}) \text{ and } p \notin U_S$, we compute $k$ nearest points in $U_S$, and do inverse distance weighted interpolation for point p. Mathematically, interpolated feature $e_p$ for point $p \in \texttt{interior}(P_{U_S}) \text{ and } p \notin U_S$ is defined as
\begin{equation}
    f_S[p] := e_p = \frac{\sum^{k}_{i=1}\frac{1}{d_i}e_{v_i}}{\sum^{k}_{i=1} \frac{1}{d_i}} \;,
\end{equation}
where $\{v_i: i \in \{1, ..., k\}, v_i \in U_S\} \subset U_S$ is the set of $k$ nearest projected vertices and $d_i = ||p-v_i||_2$. For points $p \in S \text{ and } p \notin \texttt{interior}(P_{U_S})$, we assign an feature $e_b$. This results in a dense constructed feature $f_S \in \mathbb{R}^{H \times W \times D}$ from assigned features at each point in $S$. 
We further process these features using a CNN-based encoder $E_{proc-1}$.
\begin{equation}
    f_{proc-1} = E_{proc-1}(f_S) \;,
\end{equation}
where $f_{proc-1} \in \mathbb{R}^{H \times W \times D_{out}}$. 

We add these features to the source image's dense DINOv2 features $f^{src}_{dense} \in \mathbb{R}^{H \times W \times D_{out}}$ and process the result with another CNN-based encoder $E_{proc-2}$. Mathematically, the output of our \ModuleName $f_{reg}$ is 
\begin{equation}
    f_{reg} = E_{proc-2} (f^{src}_{dense}+f_{proc-1}) \;.
\end{equation}

In our implementation, we use $H = W = 296$ and $D_{out}=256$ as in \cite{chu2024gagavatar}. We set $k=11$ and $D=512$ for all our comparisons.  

\textbf{Objective Functions.}
In order to make sure that the \ModuleName learns meaningful features, we constrain the training with two losses. First, 
we use the MSE loss between the driving image's DINOv2 features ($f^{dri}_{dense}$) and output of the \ModuleName $f_{reg}$. 
\begin{equation}
    L_{feat} = ||f^{dri}_{dense} - f_{reg}||^2_{2} \;.
\end{equation}
Next, to ensure that the features learned in the \ModuleName are similar to each other, we regularize the embeddings $e$. This is enforced by 
\begin{equation}
    L_{reg} = \frac{\texttt{pcos}(e)}{n(V)(n(V)-1)} \;,
\end{equation}
where $\texttt{pcos}(X) = \sum_i \sum_j \frac{X_i \cdot X_j}{||X_i||||X_j||} - n(V)$ is the sum of the non-diagonal elements of a self-pairwise cosine distance.
We use a weighted combination of $L_{feat}$ and $L_{reg}$ with weights $\lambda_{feat}$ and $\lambda_{reg}$. Together, they form $L_{register} = \lambda_{feat}L_{feat}+\lambda_{reg}L_{reg}$.
In our implementation, we set $\lambda_{reg}=20$ and $\lambda_{feat}=2$.

\textbf{Avatar Adaptation and Generation.}
To adapt to a particular identity, we pick the first frame of a video as the source image and select a random frame as the driving image. Next, we predict the \ModuleName's output $f_{reg}$ from the source image's DINOv2 features $f^{src}_{dense}$ and driving image's 3DMM parameters. Next, we pass the driving image's 3DMM parameters to the expression branch and $f_{reg}$ to the reconstruction branch. The outputs of the corresponding branches are merged to produce a coarse image. A neural renderer then produces the fine image. During inference, we skip the \ModuleName and directly pass the source image's DINOv2 features $f^{src}_{dense}$ to the reconstruction branch, while the rest of the process is the same as the training stage.

\section{Experiments}
\label{sec:experiments}

\subsection{Dataset Collection}

We propose a new dataset, namely RareFace-50. Prior work uses datasets with a large number of identities, e.g.~VFHQ~\citep{xie2022vfhq}. However, these datasets mostly include videos of celebrities and well-known faces from television. Thus, they might lack in diversity in terms of age and high-frequency facial details, such as wrinkles or unique tattoos (see Figure~\ref{fig:ablat_visual}). These underrepresented human faces are difficult to faithfully generate by generic networks, such as GAGAvatar~\citep{chu2024gagavatar}. Identifying this issue in existing datasets, we collect a video dataset of 50 identities with \textit{unique facial details} from YouTube. The dataset is collected from high-resolution close-up videos shot in 1080p, 2K and 4K formats. We detect faces, crop and resize face images to $512\times 512$ resolution. The average duration of the videos in this dataset is around 15 seconds, with 2 videos per identity, resulting in the total number of videos in the dataset equal to 100. We intend to publish the dataset for research purposes.
In addition to RareFace-50, we also use VFHQ test set to evaluate our method. VFHQ Test consists of 50 high quality videos from 50 different identities cropped and resized to $512\times 512$ resolution. Each video is around 4 to 10 seconds in diverse poses and settings.

We pre-process input videos using the tracking pipeline from \cite{chu2024gagavatar}. This step provides background-matted input frames, along with its tracked 3DMM parameters (these include view pose, eye pose, jaw pose, FLAME shape and expression parameters). We also pre-compute visible vertices of the 3DMM mesh fitted on a particular frame. After this, we also compute the alpha-shape polygon for the projection of the visible vertices and the set of points that lie within this polygon given the scale of the projection screen size. See more details in the supplementary material.

\subsection{Learned Features}
\begin{figure}[!t]
    \centering
    \includegraphics[width=\linewidth]{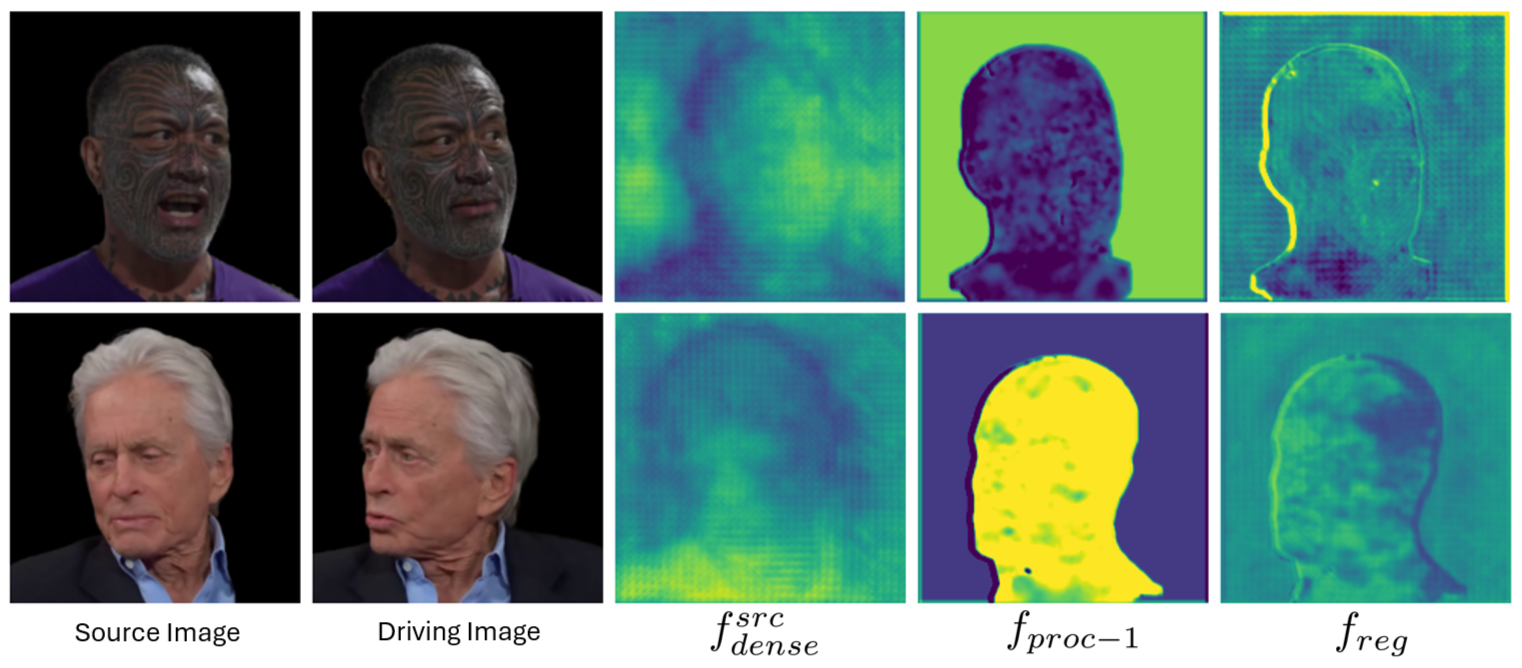}
    \caption{Visualization of learned features by the \ModuleName on our RareFace-50 dataset.
    We visualize 1) source image's DINOv2 feature $f^{src}_{dense}$, 2) $f_{proc-1}$, output from $E_{proc-1}$, and 3) $f_{reg}$, output of the \ModuleName.
     We compute the 2nd channel-wise PCA component and standardize the values.
    We observe that the \ModuleName improves the learning signals by highlighting face regions and dampening the background regions.}
    \label{fig:ablat_visual}
        \vspace{-5pt}
\end{figure}
In Figure~\ref{fig:ablat_visual}, we visualize the learned features of our \ModuleName. Specifically, we visualize features 1) $f_{proc-1}$ from $E_{proc-1}$, and 2) $f_{reg}$ from $E_{proc-2}$. We compute the 2nd channel-wise PCA component of DINOv2 features, and standardize and visualize using colormap between the range $[-3\sigma, 3\sigma]$ for DINOv2 features and $[-\sigma, \sigma]$ for $f_{proc-1}$ and $f_{reg}$. Since the DINOv2 model is trained in a self-supervised manner to make features of different augmented views of an input image to be similar on a diverse dataset, we observe features predicted by DINOv2 to have features \textit{irrelevant} to the task at hand, i.e., representing human faces. Moreover, the addition of the features from $f_{proc-1}$ changes the characteristics of the added DINOv2 features (see the comparison of $f^{src}_{dense}$ and $f_{reg}$ in Figure \ref{fig:ablat_visual}), changing the distribution in irrelevant regions of the DINOv2 features. In the supplementary material, we show additional visualization results, other PCA components, and an analysis on the norms of DINOv2 and \ModuleName features indicating that they improve meaningfully.

\subsection{Ablation Study}
\begin{figure}[!t]
    \centering
    \includegraphics[width=\linewidth]{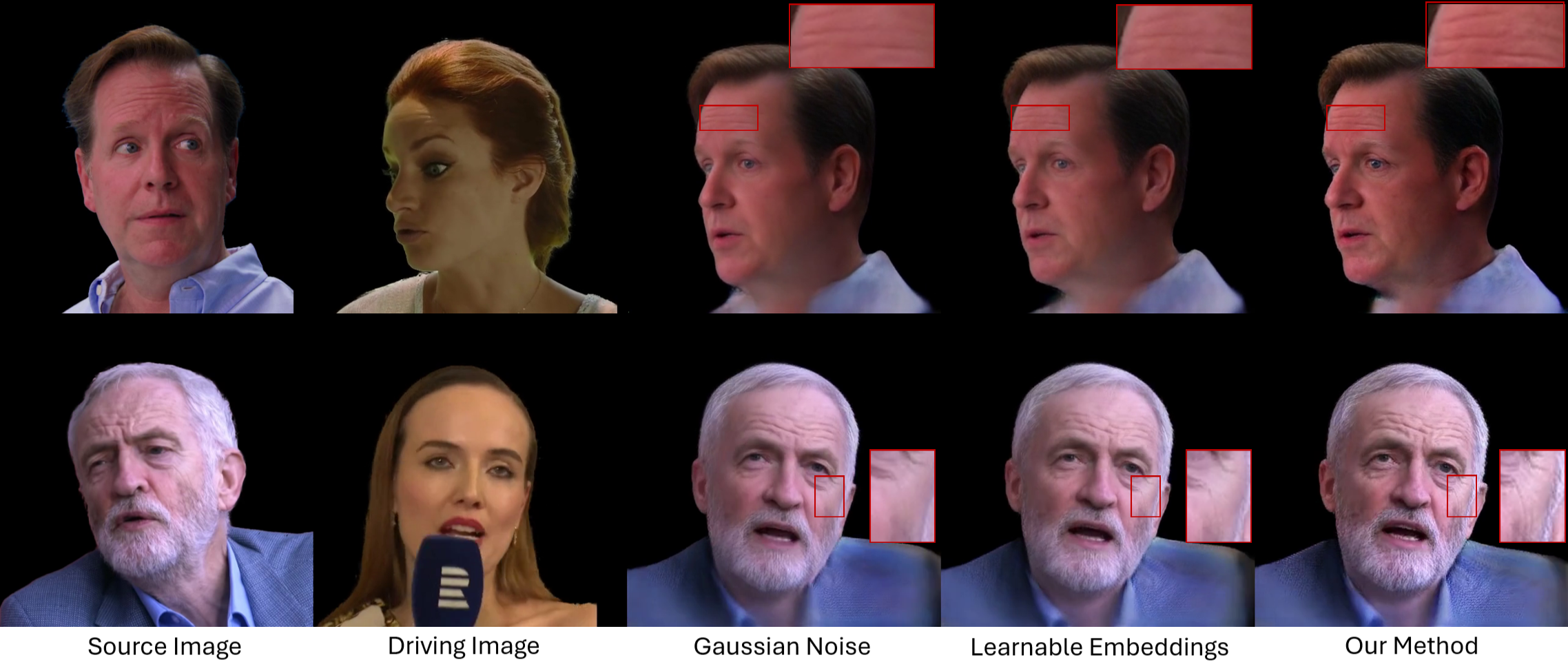}
    \caption{Ablation results on the VFHQ Test dataset. We observe that our method performs better in preserving fine-details such as wrinkles and blemishes.}
    \label{fig:ablat}
\end{figure}

\begin{table}[!t]
  \caption{Ablation study on our proposed \ModuleName. In (a), we add noise to DINOv2 features during adaptation. In (b), we add learnable embeddings to DINOv2 features during adaptation.
  }
  \label{tab:ablation}
  \centering
  \small
  \begin{tabular}{l|cccc}
    \toprule
    Method     & LPIPS$\downarrow$ & ACD$\downarrow$ \\
    \midrule
    (a) Gaussian Noise & 0.2699 & 0.3795\\
    (b) Learnable Embeddings  & 0.2622 & 0.3604\\
    Ours & \textbf{0.2470} & \textbf{0.3559}\\
    \bottomrule
  \end{tabular}
\end{table}

We conduct an ablation study on our \ModuleName comparing our method with variants. This study shows the contribution of our proposed 3D feature space, which extends the idea of registers to 3D human faces.
Specifically, in variant (a) Gaussian Noise, we sample gaussian noise $G = (G_{h, w, d}) \in \mathbb{R}^{H \times W \times D_{out}}$, $Z_{h,w,d} \overset{\mathrm{iid}}{\sim} \mathcal{N}(0,1) 
$ and add this noise $G$ to the features $f^{src}_{dense}$ during the adaptation stage. During inference, we directly use $f^{src}_{dense}$. In variant (b) Learnable Embeddings, we add a learnable embedding dictionary $e_{learn} \in \mathbb{R}^{H \times W \times D_{out}}$ to $f^{src}_{dense}$ and make it trainable during adaptation. Again, during inference we directly use $f^{src}_{dense}$.

Figure~\ref{fig:ablat} shows visual results from these variants. Adding Gaussian Noise causes washed out colors and overly smoothed details. Adding Learnable Embeddings improves the preservation of details and colors slightly. This variant would be the immediate extension of registers from~\cite{darcet2023vision} to our case. However, we notice that our proposed \ModuleName best preserves high-frequency details, such as roughness of the face and fine wrinkles, by learning an appropriate 3D feature space for human faces.
Table~\ref{tab:ablation} shows the corresponding quantitative results, demonstrating the efficacy of our module in enhancing identity-specific details. We see that our method has better perceptual similarity to the input image and better preserves identity.
We encourage the readers to watch our supplementary video for additional results demonstrating the efficacy of our \ModuleName.
\subsection{Evaluation}

\noindent
\textbf{Baselines.}
We compare our method to a baseline as the generic avatar generation model \citep{chu2024gagavatar} and the state-of-the-art approaches, namely LoRA~\citep{saunders2024talklora} and MetaPortrait~\citep{zhang2023metaportrait} for adaptation in the cross-reconstruction setting. We use the same rank $r=32$ for all our comparisons. 
Note that we implement the meta-learning algorithm from MetaPortrait~\citep{zhang2023metaportrait} on LoRA weights for fair comparisons.

\noindent
\textbf{Evaluation Metrics.}
To measure visual quality, we select challenging patches with high-frequency details from predicted frames and compare them against source image patches using 
Learned Perceptual Image Patch Similarity (LPIPS)~\citep{lpips}
in the cross-reconstruction setting. 
Furthermore, we estimate the identity preservation using the Average Content Distance (ACD) metric~\cite{sda}, by calculating the cosine distance between ArcFace~\citep{deng2019arcface} face recognition embeddings of synthesized and source images. Essentially, the idea is that the smaller the distance between those embeddings, the closer are the synthesized images to the input source images in terms of identity.

\begin{table}[!t]
  \caption{Quantitative comparisons of our approach with the baseline and other state-of-the-art adaptation methods. Results are highlighted as follows: \colorbox[HTML]{FF8A33}{Best} and \colorbox[HTML]{FFE993}{Second Best}.}
  \label{tab:baselines}
  \centering
  \small
  \begin{tabular}{l|cc|cc}
    \toprule
    & \multicolumn{2}{c|}{VFHQ Test} & \multicolumn{2}{c}{RareFace-50}\\
    \midrule
    Method     & LPIPS$\downarrow$ & ACD$\downarrow$ & LPIPS$\downarrow$ & ACD$\downarrow$ \\
    \midrule
    GAGAvatar~\citep{chu2024gagavatar} & \colorbox[HTML]{FFE993}{0.2540} & \colorbox[HTML]{FFE993}{0.3631} & 0.2953 & 0.4046\\
    LoRA~\citep{hu2022lora} & 0.2666 & 0.3687 & \colorbox[HTML]{FFE993}{0.2855} & \colorbox[HTML]{FFE993}{0.3872} \\
    MetaPortrait~\citep{zhang2023metaportrait} & 0.2650 & 0.4131 & 0.2901 & 0.4504\\
    Ours & \colorbox[HTML]{FF8A33}{0.2470} & \colorbox[HTML]{FF8A33}{0.3559} & \colorbox[HTML]{FF8A33}{0.2744} & \colorbox[HTML]{FF8A33}{0.3792}\\
    \bottomrule
  \end{tabular}
\end{table}

\noindent
\textbf{Quantitative Evaluation.}
Table~\ref{tab:baselines} shows our quantitative results.
Our method significantly outperforms the state-of-the-art in low-rank adaptation (LoRA and Meta Learning on LoRA) in terms of visual quality (LPIPS) and identity preservation (ACD).
We encourage the readers to watch our 
supplementary video for additional results demonstrating the efficacy of our \ModuleName.
\noindent
\textbf{Qualitative Evaluation.}
\begin{figure}[!t]
    \centering
    \includegraphics[width=\linewidth]{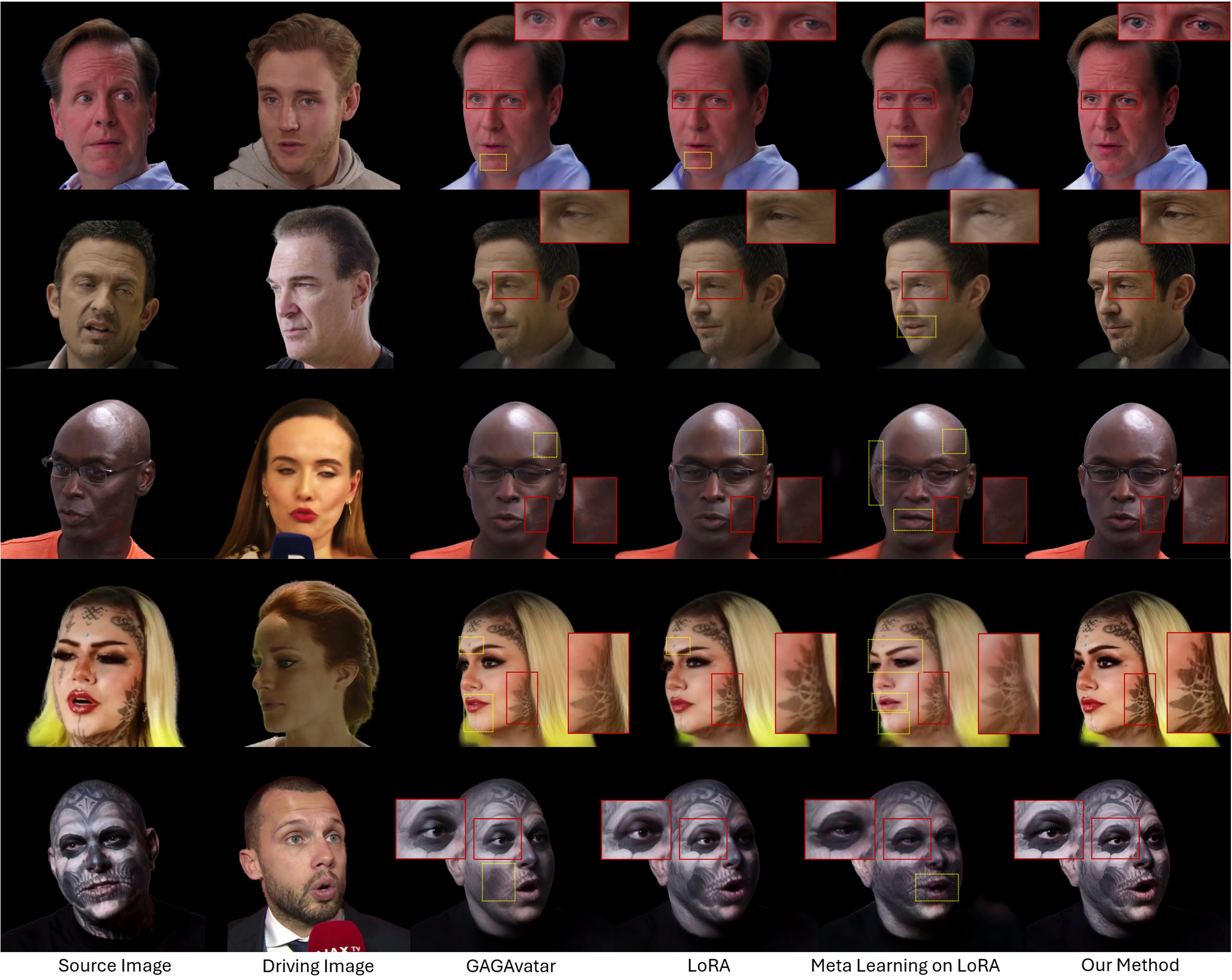}
    \caption{Personalized head avatar generation on VFHQ Test (row 1, 2, and 3) and RareFace-50 (rows 4 and 5). We compare state-of-the-art methods for adaptation (LoRA \citep{hu2022lora} and Meta-Learning \citep{zhang2023metaportrait}). We observe that our method preserves fine details for identity-specific features and produces higher quality results as compared to other methods.}
    \label{fig:main_figure}
\end{figure}\textbf{}
Figure~\ref{fig:main_figure} shows our qualitative results. Notice how our method faithfully reconstructs fine details and identity-specific features well as compared to other methods. GAGAvatar frequently produces washed out colors (in rows 4 and 5) and muted details (rows 1, 2, and 3). While LoRA performs better than GAGAvatar, it still misses fine wrinkles and veins (rows 1, 2, and 3), contrast in tattoos and skin (rows 4 and 5), and bumpy skin (row 3). Meta Learning on LoRA weights produces artifacts on face (row 3) and in eyes (rows 1, 2, and 5), produces wrong expression as compared to the driving image (all rows), misses tattoos on skin (row 4), and generates wrong colored lips. In general, our method learns to preserve high-frequency details in the source identity and produce higher quality results while using the same number of parameters as other methods.

\section{Conclusion}
\label{sec:conclusion}
In conclusion, we introduce a novel method for personalized head avatar generation. State-of-the-art approaches for adaptation such as vanilla LoRA and meta-learning fail to preserve high-frequency details and identity-specific features. We propose a novel \ModuleName that enhances the performance of LoRA, by teaching the layers to attend to specific regions in the intermediate features of a pre-trained model. To demonstrate the effectiveness of our method, we collect a dataset of talking individuals with distinctive facial features, such as wrinkles and tattoos. Our method outperforms existing methods qualitatively and quantitatively, faithfully capturing unseen identities.

\noindent
\textbf{Limitations and Future Work.}
Although our \ModuleName successfully captures distinctive facial details, it might produce suboptimal results for extreme side or back views that are rarely or not at all seen in a video. In the future, we plan to extend our work to faithfully animate avatars from such rare views of individuals.

\noindent

{
    \small
    \bibliographystyle{plain}
    \bibliography{egbib}
}


\appendix
\section*{Appendix}

The appendix is organized as follows: 
\begin{enumerate}
    \item Additional Ablation Study in Sec. \ref{sec:addnl_ablation}
    \item Implementation Details in Sec. \ref{sec:suppl_imp_details}
    \item Dataset Collection Details in Sec. \ref{sec:data_collection}
    \item Additional Results in Sec. \ref{sec:addnl_results}
    \item User Study Details in Sec. \ref{sec:user_study}
    \item Discussion: Broader Impact, Limitations, and Ethical Considerations in Sec. \ref{sec:discussion}
\end{enumerate}
We strongly encourage the readers to watch our supplementary video.

\section{Additional Ablation Study}
\label{sec:addnl_ablation}
\begin{table}[h]
  \caption{Ablation study on losses to adapt with our method. In (a), we remove $L_{feat}$ during adaptation. In (b), we remove $L_{reg}$ during adaptation.
  }
  \label{tab:ablation_losses}
  \centering
  \small
  \begin{tabular}{l|cccc}
    \toprule
    Method     & LPIPS$\downarrow$ & ACD$\downarrow$ \\
    \midrule
    (a) Ours w/o $L_{feat}$ & 0.2574 & 0.3604\\
    (b) Ours w/o $L_{reg}$  & 0.2508 & 0.3574\\
    Ours & \textbf{0.2470} & \textbf{0.3559}\\
    \bottomrule
  \end{tabular}
\end{table}
\begin{table}[h]
  \caption{Ablation study on length of videos to adapt the head avatar. We set the adaptation video length to 4, 2, and 1 seconds.
  }
  \label{tab:ablation_length}
  \centering
  \small
  \begin{tabular}{l|cccc}
    \toprule
    Method     & LPIPS$\downarrow$ & ACD$\downarrow$ \\
    \midrule
    (a) Ours w/ 4 sec & \textbf{0.2471} & \textbf{0.3639}\\
    (b) Ours w/ 2 sec  & 0.2471 & 0.3656\\
    (c) Ours w/ 1 sec & 0.2477 & 0.3687\\
    \bottomrule
  \end{tabular}
\end{table}

We conduct additional ablation studies on our proposed method. Specifically, we ablate the various losses we propose. In Table~\ref{tab:ablation_losses}(a), we remove the loss $L_{feat}$ to supervise the output of our register module during adaptation, and in (b) we remove $L_{reg}$ to make the learned embeddings in our \ModuleName different from each other. We see that removal of these losses causes a drop in performance as compared to when both losses are present. Furthermore, we ablate on the length of the videos used to adapt our head avatars in Table~\ref{tab:ablation_length}. We see that reducing the length of the adaptation video causes a drop in the performance of the method. Note that we trim the original videos to the first 4, 2, and 1 second for these experiments.
 
\section{Implementation Details}
\label{sec:suppl_imp_details}

\subsection{Addition of LoRA to layers}
\label{sec:description_lora}
We use minLoRA~\citep{minLoRA} library to add LoRA~\citep{hu2022lora} parameters to all layers of a pretrained pytorch model. 
During training LoRA is instantialized as separate parameters $B \in \mathbb{R}^{m \times r}, A \in \mathbb{R}^{r \times n}$ and $r \ll min(m, n)$ from the pretrained parameters $W_{pre} \in \mathbb{R}^{m \times n}$ of the module, so that these parameters can be trained. During inference the LoRA parameters are \textit{merged} with the pretrained parameters and assigned as the new pretrained parameters according to:
\begin{equation}
    W_{adapt} := W_{pre} + \Delta W = W_{pre} + BA.
\end{equation}
This ensures that the LoRA layers add no overhead to the pipeline during the inference. 
Given a pretrained GAGAvatar model, we add LoRA weights to all layers except in the DINOv2 feature extractor. 

\subsection{Dataset Preprocessing}
\label{sec:data_preprocessing}
Given the expression code, shape code, and camera pose predicted during 3DMM fitting, we predict a 3DMM mesh. We compute visible vertices of 3DMM mesh from the computed camera pose of a particular frame using trimesh's RayMeshIntersector implementation. Specifically, we cast rays from the camera origin to each point in the mesh and compute whether any line intersects the mesh (if an intersection exists, the point is not visible). Given these visible points in 3D space, we find a screen space camera projection on a screen of the same size as DINOv2 feature space ($H = W = 296$) using Perspective Projection. Then we find an alpha-shape of these projected points with $\alpha = 0.065$. Next, we compute all the points in the alpha-shape polygon using a parallelized point-in-polygon test. For all points in the polygon that are not projected points, we also compute the $k$ nearest projected points and distances from those projected points, where $k=11$. 

\subsection{Register Module Details}
\label{sec:register_module_details}
The embeddings rigged to vertices on the 3DMM mesh~\citep{FLAME:SiggraphAsia2017} are modeled in a 3D space in our register module, However, these points are projected onto a 2D space using a camera projection following which, we interpolate the features using a weighted sum of the $k$ nearest neighbors to fill up the face region in densely constructed feature $f_S$. Using the entire set of the vertices would cause all points to be projected onto the densely constructed feature $f_S$, thus impacting the interpolation process (projection of points from the back of the head might be in the $k$ nearest neighbors of a point $p \in interior(P_{U_S})$). Thus, we only project visible points from any given view in our register module. 

We model $E_{proc-1}$ as a convolutional module, with 4 convolutional layers of channel sizes $[512, 512, 256, 256]$ and kernel sizes set to $3$ for each layer. $E_{proc-2}$ is also a convolutional module with 4 layers of channel sizes set to 256. The first 3 layers have a kernel size of 3, while the last layer has a kernel size of 1.

\subsection{Training Details}
\label{sec:training_details}
\subsubsection{Our Method}
\label{sec:training_our_method}
We initialize embeddings $e$ using Xavier Normal initialization~\citep{pmlr-v9-glorot10a}. We adapt head avatars with our method for a total of $1000$ iterations. The batch size is set to $2$. We use Adam~\citep{kingma2017adammethodstochasticoptimization} optimizer with learning rate set to $1e-4$ for the LoRA layers whereas the learning rate is set to $1e-3$ for parameters in the \ModuleName. We use a linear learning rate scheduler with a start factor of 1.0 and an end factor of 0.1 at the $1000$th iteration. Along with our proposed losses, we also keep the losses proposed by GAGAvatar~\cite{chu2024gagavatar}, namely, RGB losses between predicted image and driving image, a perceptual loss between the predicted images and driver image, and $L_{lifting}$, loss between predicted points from reconstruction branch and vertices of the 3DMM mesh fitted on the driving image. Our adaptation takes $\approx35$ minutes on an RTX A5000 GPU, consuming $\approx23$GB of VRAM. During inference, we load and merge the LoRA weights into their corresponding layer parameters. Thus, there is no overhead during inference, i.e., it consumes the same amount of resources as GAGAvatar during inference.  

\subsubsection{Baselines and Ablations}
\label{sec:training_baselines_and_ablations}
For all comparisons with the vanilla LoRA, we use the same hyperparameters as our method. That is, we adapt with vanilla LoRA for a total of $1000$ iterations, set the batch size to $2$, use Adam optimizer with learning rate set to $1e-4$ for the LoRA layers, and use a linear learning rate scheduler with a start factor of 1.0 and an end factor of 0.1 at the $1000$th iteration. This adaptation takes $\approx25$ minutes on an RTX A5000 GPU, consuming $\approx14.9$GB of VRAM. 

Following MetaPortrait~\citep{zhang2023metaportrait}, we implement Reptile~\cite{nichol2018reptile}, a MAML based strategy for meta-learning in our low-rank adaptation task. We perform pre-training on our RareFace-50 dataset. Following the formulation of MetaPortrait, we formulate the task of adapting to a particular identity as an inner loop task. Thus, we adapt to a randomly sampled identity at each outer step. For all comparisons with Meta-Learning on LoRA, we set rank $r=32$, the inner loop learning rate to $2e-4$, and the outer update step size to $2e-5$. The number of inner loop steps are set to 120, and number of elements in batch in inner loop is set to 4. We set the number of outer iterations to 4800. Given resource constraints, we implement a single GPU version of Reptile~\citep{nichol2018reptile}, thus taking 12 days to complete the pretraining task on a Quadro RTX 8000 GPU, consuming $\approx45$GB of VRAM. After the pre-training task, we adapt the model on an identity for 120 steps with the same learning rate as the inner loop, which takes $\approx4$ minutes on an RTX A5000 GPU consuming $\approx14.9$GB of VRAM. We then use these adapted weights for inference by merging these LoRA weights to the corresponding layers. 

For all experiments with learnable parameters $e_{learn}$ as a replacement to our \ModuleName, we set the learning rate for $e_{learn}$ parameters to be $1e-3$.
\subsection{Metrics}
\label{sec:metrics_details}
We compute the visual quality metrics namely, LPIPS~\cite{lpips}, on specific challenging crops with high frequency details from predicted frames and compare them against source image patches. The identity preservation metric (ACD) is measured using ArcFace~\cite{deng2019arcface}, a ResNet50-based network trained on WebFace~\cite{zhu2021webface260m}. Specifically, we used ``buffalo\_l'' model from the insightface repository~\cite{insightface}.
\section{Data Collection}
\label{sec:data_collection}
We collect data from Youtube of people knowingly appearing in interviews in public broadcasts with distinctive facial details, such as wrinkles or tattoos. These characteristics are under-represented in existing datasets. We will present the dataset as a set of links, along with trim times and crop position coordinates. Additionally, this dataset will be maintained using an automatic script that checks and removes links from the list that no longer exist in YouTube.

\section{Additional Results}
\label{sec:addnl_results}

\subsection{Details of Adaptation}
\textbf{Adaptation Duration.}
In this section, we discuss the adaptation durations of our baselines and our method. Vanilla LoRA and our method take $\approx 25$ minutes and $\approx35$ minutes to adapt respectively on an RTX A5000 GPU. Whereas, meta-learning on LoRA requires a much longer $12$ day period on an RTX Quadro 8000 GPU for the pretraining objective, after which it requires $\approx 4$ minutes on a RTX A5000 GPU. However, during inference, all of these methods have the same inference times as GAGAvatar~\cite{chu2024gagavatar}.

\textbf{Adaptation Parameters.}
In this section, we discuss the number of parameters during adaptation and inference of our baselines and our method. During adaptation, we introduce $4.7$M parameters as LoRA weights to the pretrained layers in all baselines. Our \ModuleName adds another $18.5$M parameters. Thus, our method has $23.2$M parameters during adaptation, which is $\approx11\%$ of the total number of parameters ($199$M parameters) in GAGAvatar. During adaptation, we discard the trained \ModuleName, which lends to the same efficiency as GAGAvatar and other baselines. 

\subsection{User Study}
\label{sec:sub_user_study}
\begin{figure}[h]
    \centering
    \includegraphics[width=0.95\linewidth]{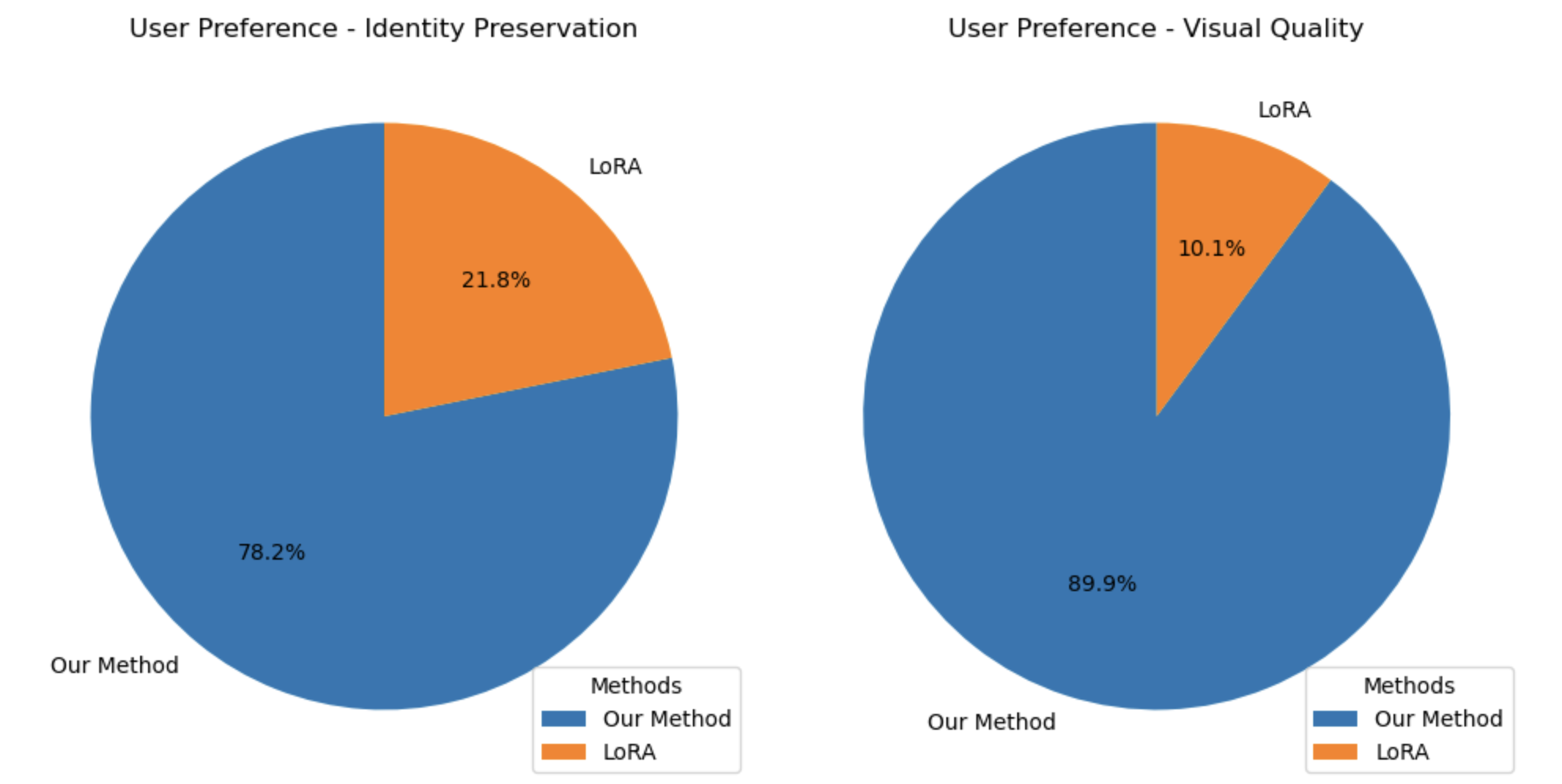}
    \caption{\textbf{User Study.} Preference (\%) in terms of identity preservation, and visual quality, comparing LoRA~\citep{hu2022lora}, and our method.}
    \label{fig:user_study_pie}
\end{figure}

We conduct a user study to qualitatively and subjectively compare our method against LoRA (see Sec.~\ref{sec:user_study} for details). The results of our user study are shown in Fig.~\ref{fig:user_study_pie}. We find that $78.2\%$ of the users prefer our method as compared to LoRA in terms of identity preservation. Furthermore, we find that $89.9\%$ of the users prefer our results as compared to LoRA in terms of visual quality. 

\subsection{Additional Qualitative Results}
\begin{figure}[!t]
    \centering
    \includegraphics[width=\linewidth]{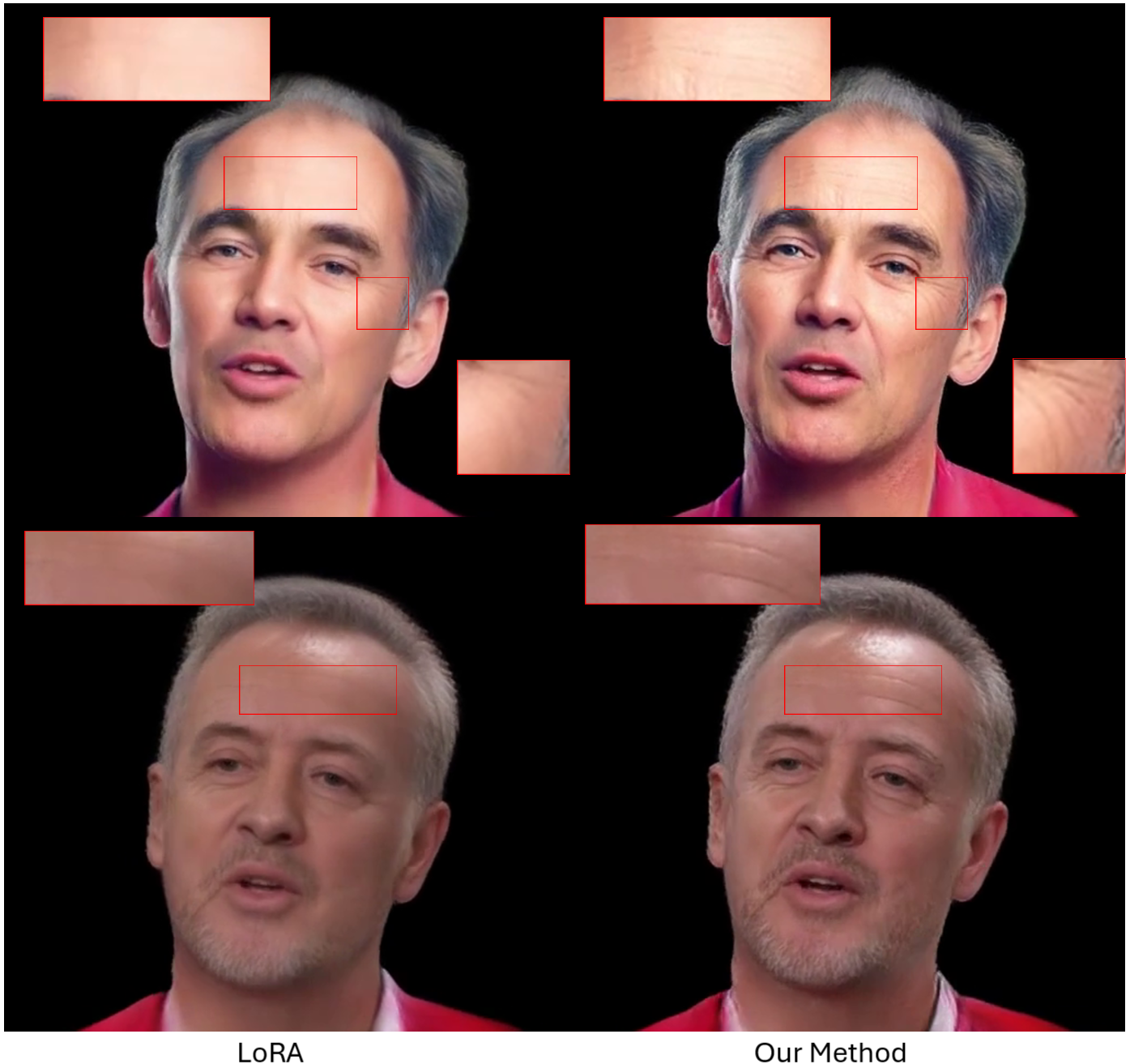}
    \caption{We show facial details that are well captured by our method with \ModuleName, such as wrinkles and skin folds are realistic and have higher quality than vanilla LoRA. Please note the enlarged insets of specific details.}
    \label{fig:high_res_details}
\end{figure}
\begin{figure}[!t]
    \centering
    \includegraphics[width=\linewidth]{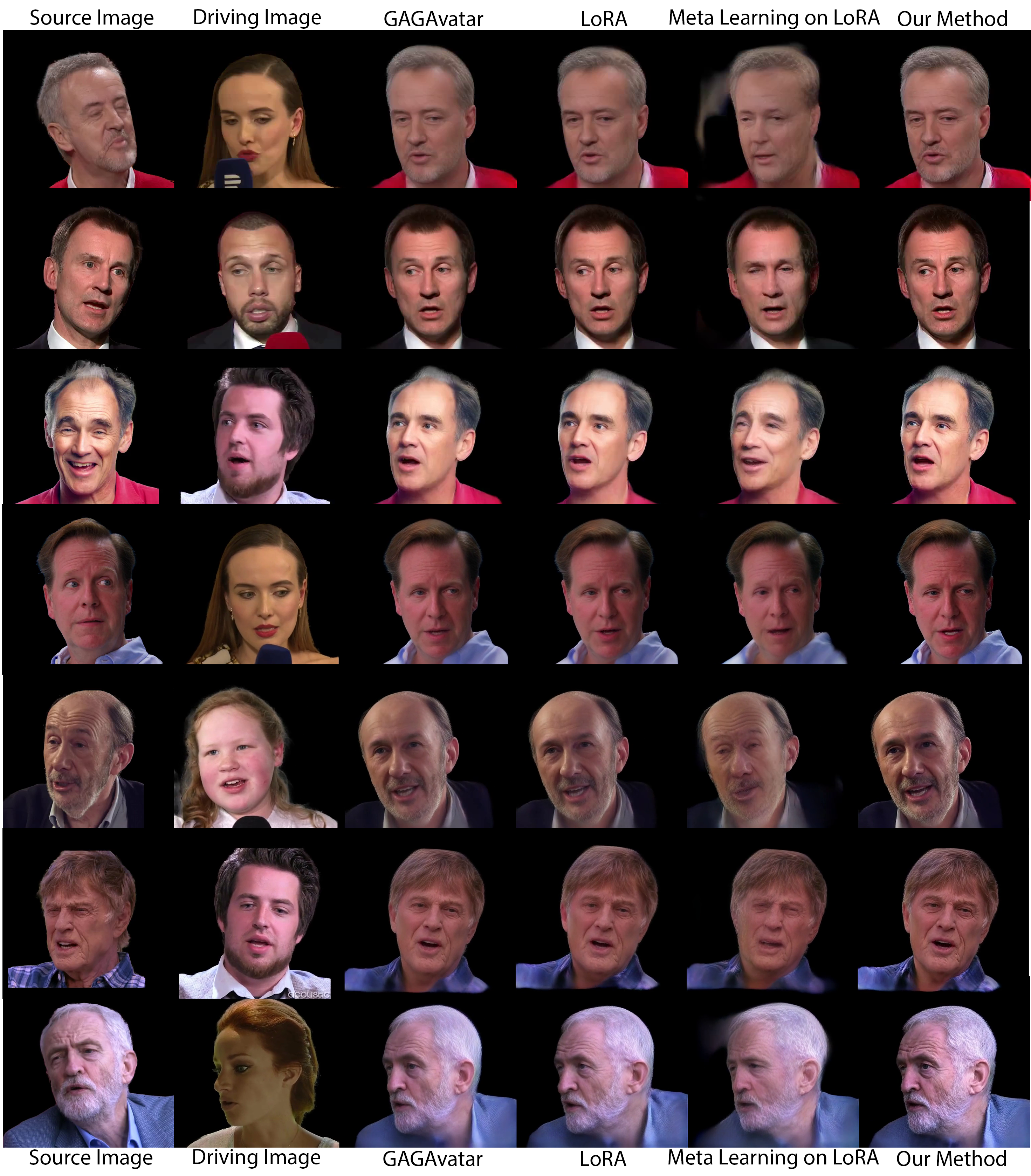}
    \caption{Additional results of personalized head avatar generation on VFHQ Test. Please zoom in for better details.}
    \label{fig:addnl_vfhq}
\end{figure}

\begin{figure}[!t]
    \centering
    \includegraphics[width=\linewidth]{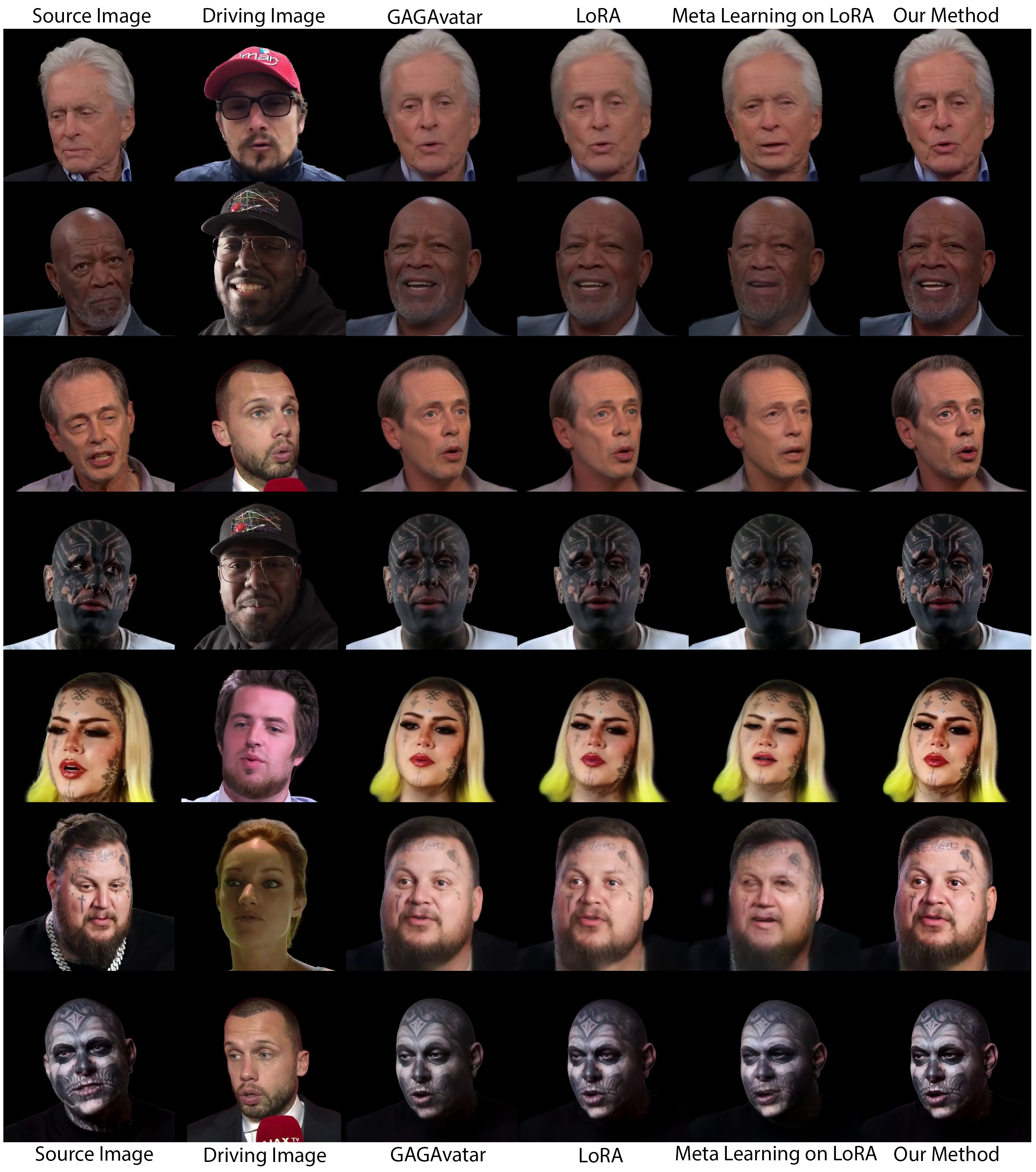}
    \caption{Additional results of personalized head avatar generation on our RareFace-50 dataset. Please zoom in for better details.}
    \label{fig:addnl_rare_face}
\end{figure}

\begin{figure}[!t]
    \centering
    \includegraphics[width=\linewidth]{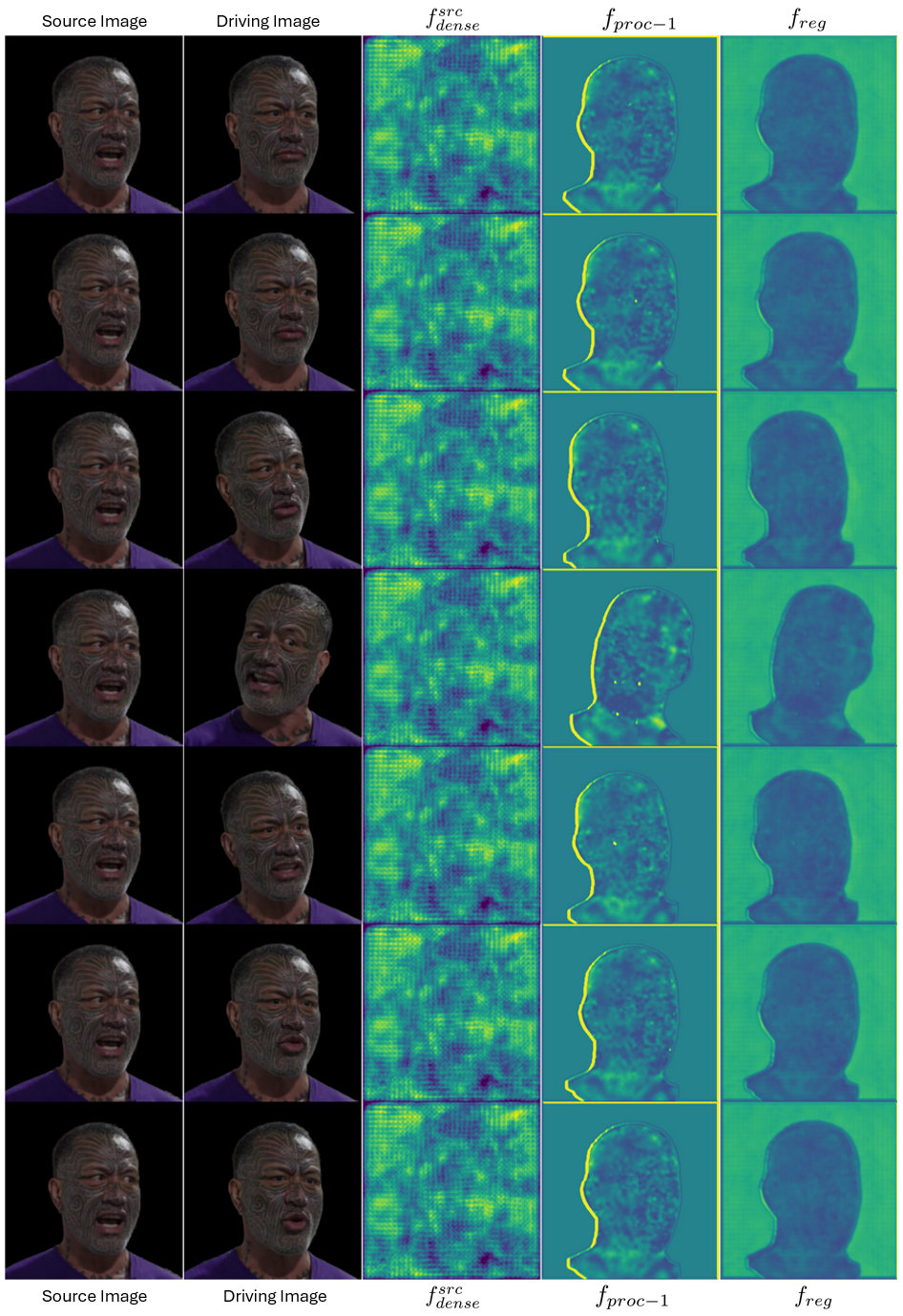}
    \caption{Visualization of learned features by the \ModuleName on our RareFace-50 dataset. We visualize the 1) source image's DINOv2 feature $f^{src}_{dense}$, 2) $f_{proc-1}$, output from $E_{proc-1}$, and 3) $f_{reg}$, output of the \ModuleName. We compute norms of the features along the embedding dimensions and standardize values. }
    \label{fig:norm_viz_ft}
\end{figure}
\begin{figure}[!t]
    \centering
    \includegraphics[width=\linewidth]{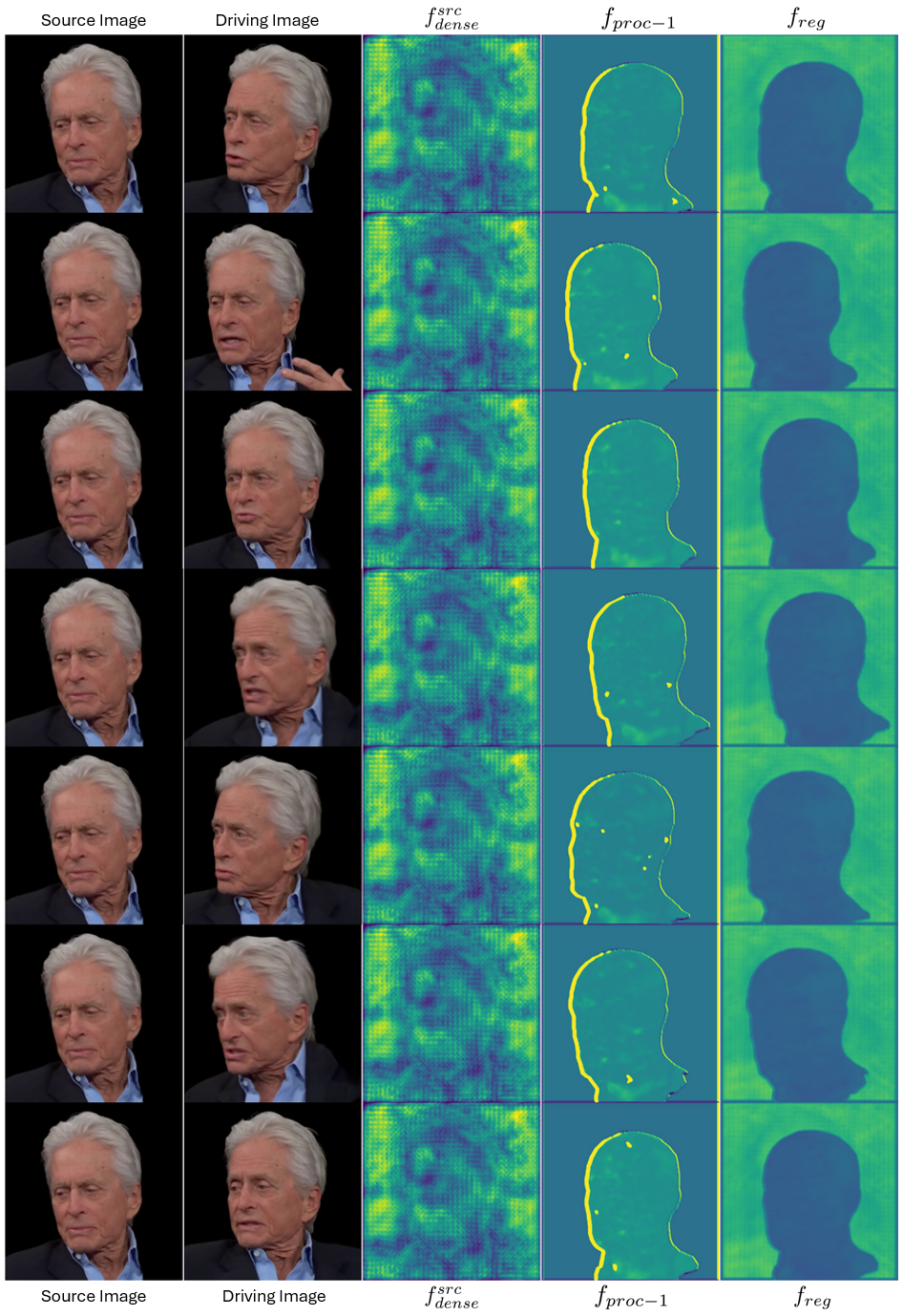}
    \caption{Visualization of learned features by the \ModuleName on our RareFace-50 dataset. We visualize the 1) source image's DINOv2 feature $f^{src}_{dense}$, 2) $f_{proc-1}$, output from $E_{proc-1}$, and 3) $f_{reg}$, output of the \ModuleName. We compute norms of the features along the embedding dimensions and standardize values.}
    \label{fig:norm_viz_md}
\end{figure}
\begin{figure}[!t]
    \centering
    \includegraphics[width=\linewidth]{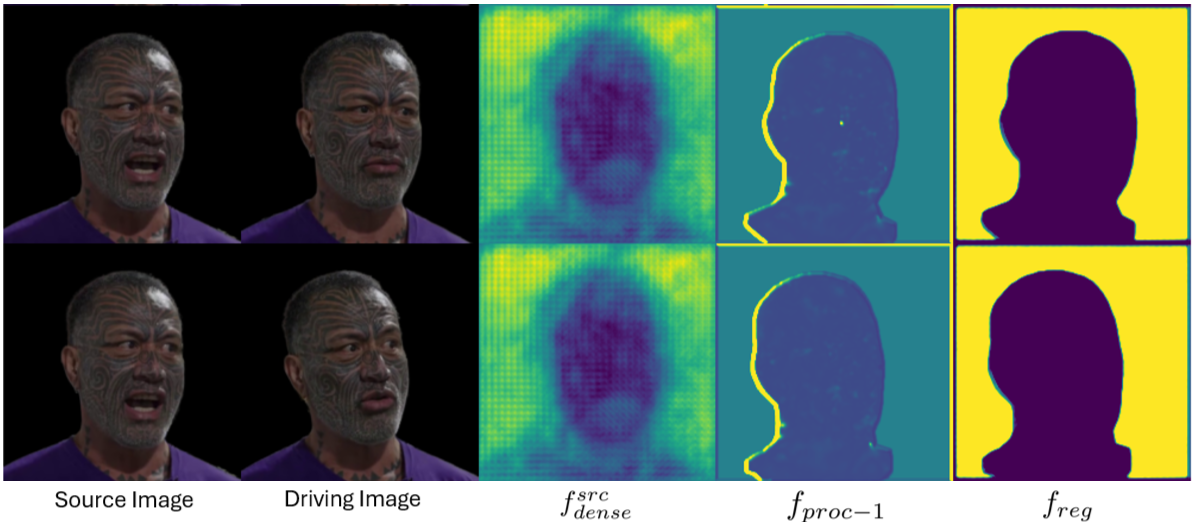}
    \vspace{-5pt}
    \caption{Visualization of learned features by the \ModuleName on our RareFace-50 dataset. We compute the 1st channel-wise PCA component and standardize the values. }
    \label{fig:pca_1_viz_ft}
    \vspace{-5pt}
\end{figure}
\begin{figure}[!t]
    \centering
    \includegraphics[width=\linewidth]{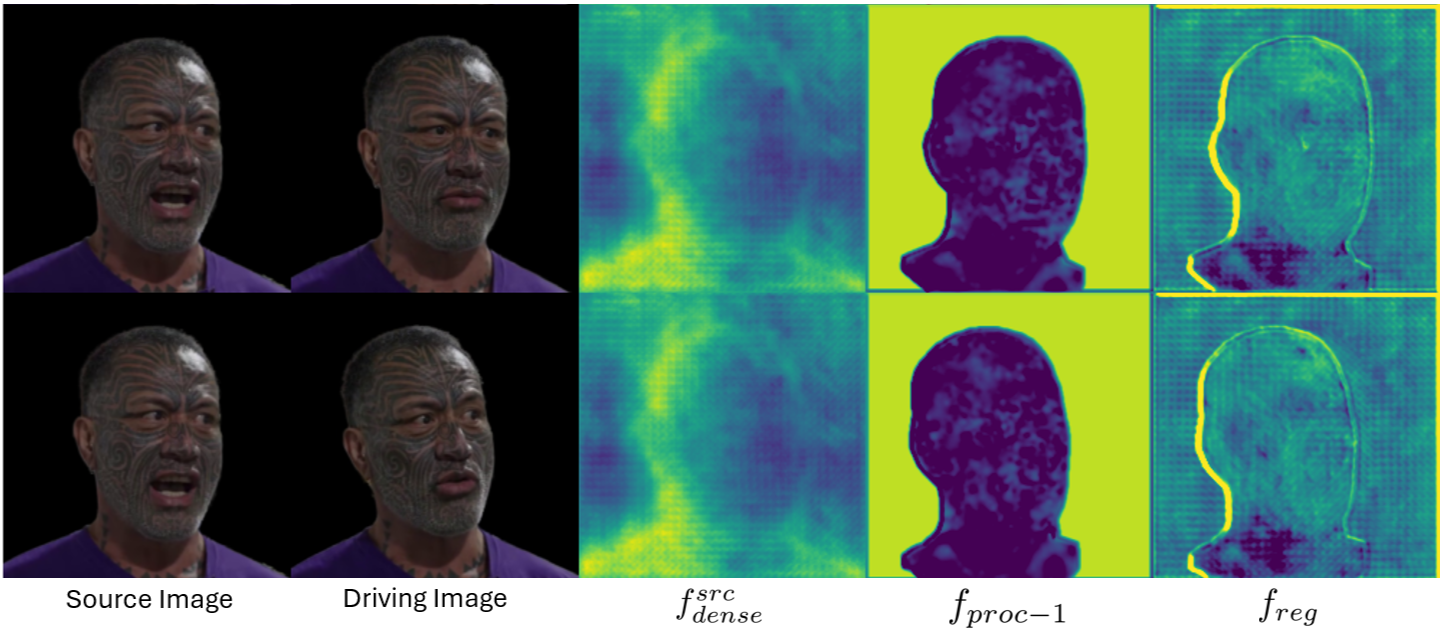}
    \vspace{-5pt}
    \caption{Visualization of learned features by the \ModuleName on our RareFace-50 dataset. We compute the 2nd channel-wise PCA component and standardize the values.}
    \label{fig:pca_2_viz_ft}
    \vspace{-5pt}
\end{figure}
\begin{figure}[!t]
    \centering
    \includegraphics[width=\linewidth]{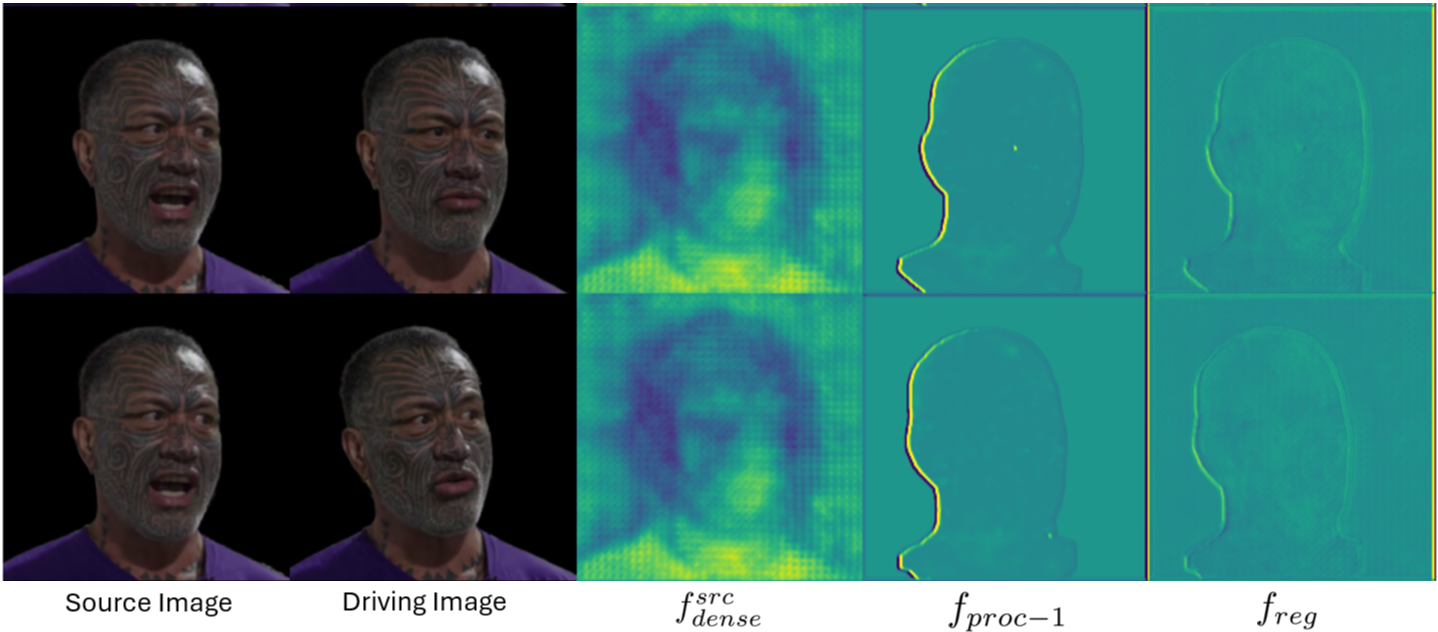}
    \vspace{-5pt}
    \caption{Visualization of learned features by the \ModuleName on our RareFace-50 dataset. We compute the 3rd channel-wise PCA component and standardize the values.}
    \label{fig:pca_3_viz_ft}
\end{figure}

In Fig.~\ref{fig:high_res_details}, we show that, compared to LoRA, our method effectively captures high-frequency details like wrinkles. We show additional results of our method against the baselines on VFHQ Test in Fig.~\ref{fig:addnl_vfhq} and on RareFace-50 in Fig.~\ref{fig:addnl_rare_face}. Fig~\ref{fig:norm_viz_ft} and \ref{fig:norm_viz_md} show visualizations of feature norms along the channel dimensions for two identities from RareFaces-50. The values are visualized using colormaps between $[-3\sigma,3\sigma]$ for $f^{src}_{dense}$, and $f_{reg}$ and $[-\sigma, \sigma]$ for $f_{proc-1}$. We visualize the first four channel-wise PCA components 
in Fig.~\ref{fig:pca_1_viz_ft} to \ref{fig:pca_3_viz_ft}. We observe that the \ModuleName improves the learning signals by highlighting face regions and dampening the background regions. 

\section{User Study Details}
\label{sec:user_study}

\begin{figure}[!t]
    \centering
    \includegraphics[width=\linewidth]{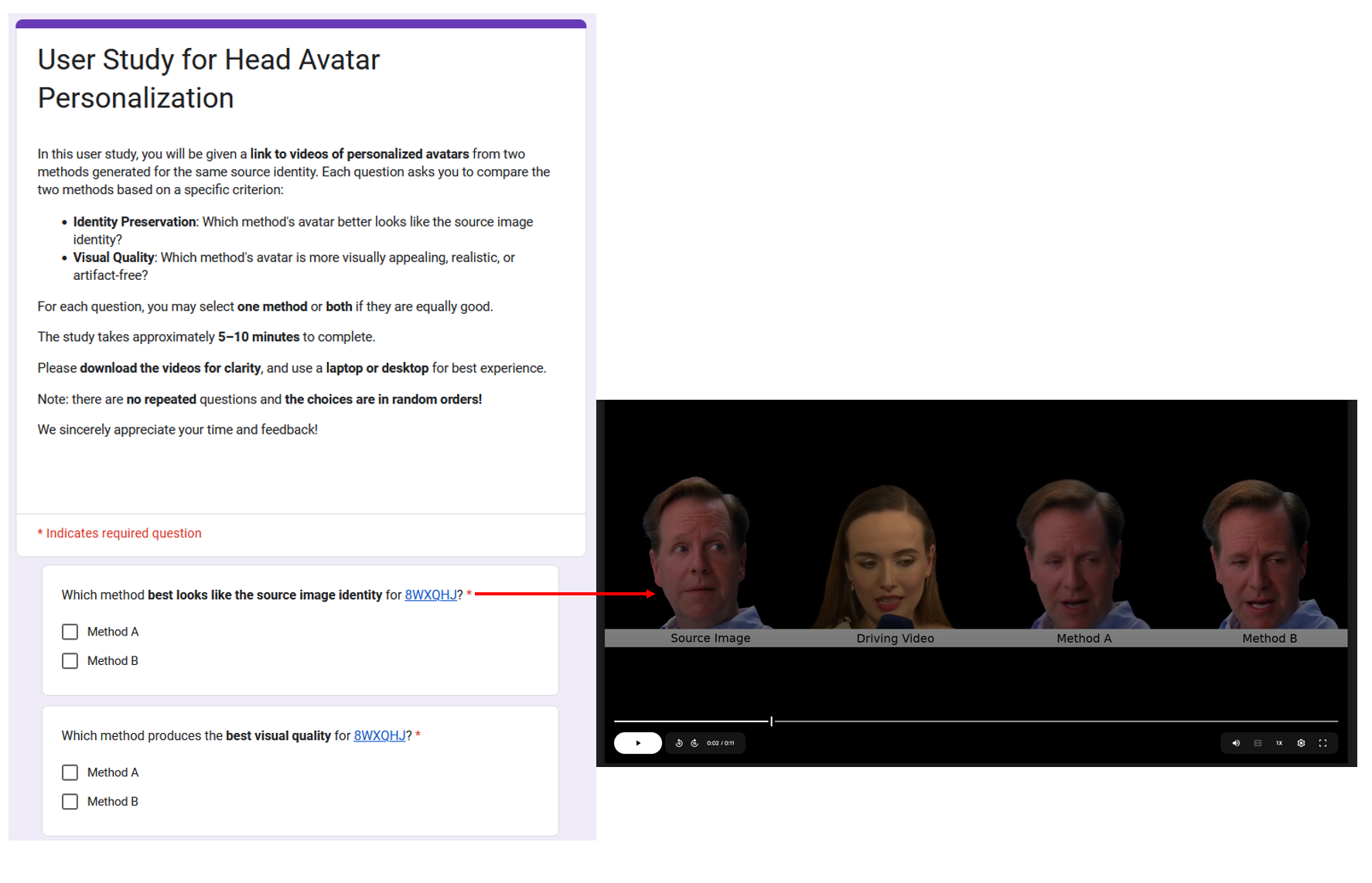}
    \caption{User Study Interface. We ask each user to watch 8 videos and answer which method preserves the source image identity and which method has the best visual quality. }
    \label{fig:user_study_details}
\end{figure}

As mentioned in Sec.~\ref{sec:sub_user_study}, we qualitatively compare our method with vanilla LoRA as an adaptation method with a user study. We describe the details of the user study here. Fig.~\ref{fig:user_study_details} shows the interface that we use for this user study. A total of 18 users responded to our user study. We generate head avatars given source videos from VFHQ Test and RareFace-50 using our method and LoRA. The outputs are placed side by side and the left-right orders are assigned randomly to make sure that the users are unaware of which method is ours. Each generated video is $\approx$ 5 to 10 seconds long, concantenated with the source identity image and the driving video. Users are asked two questions: ``Which method's avatar best looks like the source image identity?'' and ``Which method avatar has better visual quality?'' The users can choose as answer ``Method A'' or ``Method B'' or both. Label ``Method A'' is placed to the left of label ``Method B'' and the generated videos are randomly placed in terms of a left-right order. The answers are collected through a google form. The videos are attached to the google form using a link to google drive, and the users are encouraged to download the videos to view them on their system. This is done to make sure that differences in high resolution are evident to the users. 
\section{Discussions}
\label{sec:discussion}
\subsection{Limitations}
An important factor of our method is the 3DMM fitting that is used to extract the head pose, camera parameters, and 3DMM mesh parameters (see Sec. 3.3 of the main paper). This fitting can be noisy and the error can be propagated to the final generated videos. Improving the face tracking further would be an interesting future work. Further, 3DMM fitting does not model asymmetric/extreme expressions (such as winking) and the movement of the tongue, which is another interesting line of work to pursue. 

\subsection{Ethical Considerations and Broader Impacts}
While our method has significant promise across diverse applications, it also carries the risk of abuse — for example, in creating ``deep fakes''. These can be used by users with malicious intent to spread misinformation. To prevent this, it is imperative to develop forensic tools to detect fake videos \cite{cai2022marlin, reiss2023detecting}. We intend to share our code, dataset and models to improve this research, in which we will release them with strict licenses that only allow usage for academic research. When used ethically and responsibly, our method can offer profound benefits across industries — from video conferencing to the entertainment sector.
In addition, we have also put appropriate procedures (see Sec.~\ref{sec:data_collection}) to ensure fair and safe use of videos from the dataset we collect.

\end{document}